\documentclass[10pt,twocolumn,letterpaper]{article}
\usepackage[pagenumbers]{cvpr} 
\usepackage{graphicx}
\usepackage{amsmath}
\usepackage{amssymb}
\usepackage{booktabs}
\usepackage{multirow}
\usepackage{array}
\usepackage{rotating}
\usepackage{mathtools}
\usepackage{enumitem}
\usepackage[symbol,hang,flushmargin]{footmisc} 
\usepackage[pagebackref,breaklinks,colorlinks]{hyperref}
\usepackage[capitalize]{cleveref}
\crefname{section}{Sec.}{Secs.}
\Crefname{section}{Section}{Sections}
\Crefname{table}{Table}{Tables}
\crefname{table}{Tab.}{Tabs.}
\newcommand{\quotes}[1]{``#1''}
\newcommand{\klens}{K\textbar Lens GmbH, Germany}
\newcommand{\uni}{Saarland Informatics Campus, Germany}
\newcommand{\dfki}{German Research Center for Artificial Intelligence (DFKI), Germany}
\newcommand{\mpi}{MPI Informatics, Germany}

\makeatletter
\def\thanks#1{\protected@xdef\@thanks{\@thanks
		\protect\footnotetext{#1}}}

\begin{document}
	\title{Edge-aware Consistent Stereo Video Depth Estimation}
	\author{Elena Kosheleva$^{1,2}$$^{,}$* \hspace{1cm} Sunil Jaiswal$^{1,}$* \hspace{1cm} Faranak Shamsafar$^{1}$, \\
		Noshaba Cheema$^{2,3,4}$ \hspace{1cm} Klaus Illgner-Fehns$^{1}$ \hspace{1cm} Philipp Slusallek$^{2,3}$
		\vspace{0.5cm} \\
		$^{1}$ \klens \\
		$^{2}$ \uni \\
		$^{3}$ \dfki \\
		$^{4}$ \mpi \\
		\thanks{		\hspace{-0.6cm} This work was partially funded by the German Ministry for Education and Research (BMBF).}}
	\maketitle
	\footnotetext[1]{Equal contribution}
	\footnotetext[2]{Corresponding author: {\tt\small sunil.jaiswal@k-lens.de}}
	\begin{abstract}
		Video depth estimation is crucial in various applications, such as scene reconstruction and augmented reality. In contrast to the naive method of estimating depths from images, a more sophisticated approach uses temporal information, thereby eliminating flickering and geometrical inconsistencies. We propose a consistent method for dense video depth estimation; however, unlike the existing monocular methods, ours relates to stereo videos. This technique overcomes the limitations arising from the monocular input. As a benefit of using stereo inputs, a left-right consistency loss is introduced to improve the performance. Besides, we use SLAM-based camera pose estimation in the process. To address the problem of depth blurriness during test-time training (TTT), we present an edge-preserving loss function that improves the visibility of fine details while preserving geometrical consistency. We show that our edge-aware stereo video model can accurately estimate the dense depth maps.
	\end{abstract}
	\section{Introduction}
	Depth estimation refers to computing the distance to the camera viewpoint for each visible object in the scene. It is possible to acquire a depth map from a sensor, such as Kinect or LiDAR. However, such sensors are more expensive and less portable than RGB cameras. Furthermore, these sensors are imperfect and tend to leave holes and inconsistencies in the resulting depth maps. Therefore, it is preferable to compute depth maps directly from monocular or stereo data in many situations. With the advancements in virtual reality, augmented reality, and 3D visual effects in recent years, it is now necessary to continuously update depth maps based on video sequences.
	
	To this end, frame-by-frame estimating images' depth ignores the video's temporal consistency and produces flickers. Several attempts have been made to implicitly constrain temporal dependency in a video with Recurrent Neural Networks (RNN) \cite{eom2019temporally, tananaev2018temporally, patil2020don}. However, such approaches do not use scene geometry, and experiments on real data show that their performance is not satisfactory. To expose more constraints, other works, such as scene-flow methods aim to estimate depths and camera motion jointly \cite{deeprigid, MFuseScene, chen2019self, teed2018deepv2d, zhou2017unsupervised, vijayanarasimhan2017sfm}.  This helps in obtaining an understanding of the scene geometry. However, the disadvantage of these algorithms is that they require large datasets for training. The methods are also limited to inputs similar to the training data and fail to achieve accurate results for unfamiliar images.
	
	Geometrically consistent video depth estimation is another solution to produce temporally dependent depth maps \cite{yoon2020novel,CVDE, zhang2021consistent, RCVDE,ECVDE}. The motivation of our work is such a strategy, and similar to \cite{CVDE}, we propose a test-time training (TTT) algorithm to fine-tune an existing image-based depth estimation model and fit geometrical constraints in a specific video. Even though \cite{CVDE} produces robust and consistent results, it suffers from several problems. Firstly, monocular video data is the only possible input for this method, while stereo or multi-view inputs provide more possibilities for quality enhancement. Secondly, fine details and edge information are lost during fine-tuning, resulting in a blurred output. Lastly, the preprocessing step is highly time-consuming and prone to error in challenging sequences. 
	
	In this work, we take a leap beyond the monocular input and compute depth maps for stereo video data. The depth estimation is temporally consistent, flicker-free, and obtains a stable 3D geometry of the scene. Figure \ref{fig:blockdiagram} shows the proposed framework for test-time training on stereo videos. Our main contributions are:
	\begin{figure}[tbp]
	\centering
	\includegraphics[width=1\linewidth]{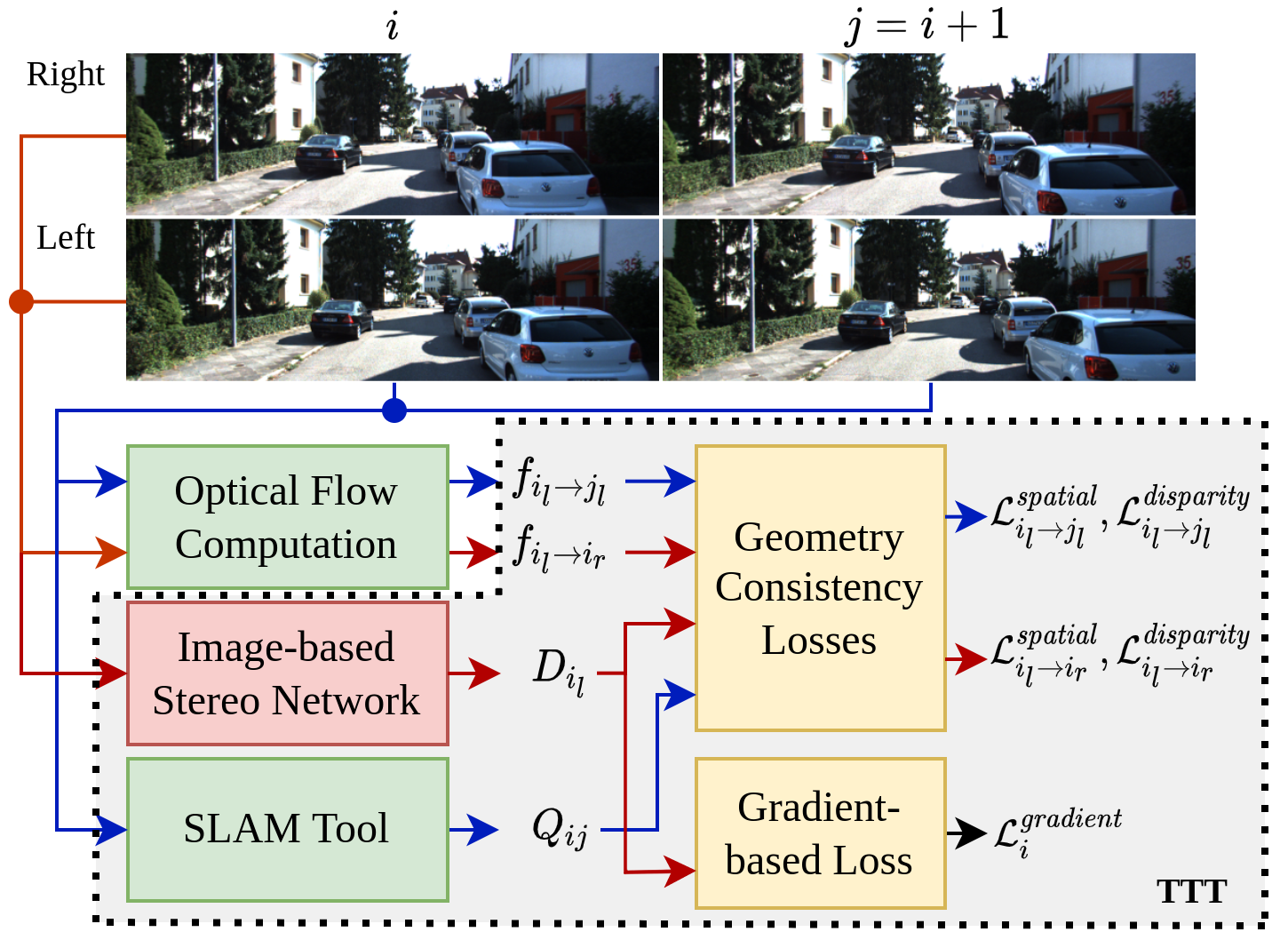}
	\caption{Overview of our test-time training (TTT) on stereo video. Here, we show the loss computations for frame $i$. Our model enforces geometric consistency between the left and right image pair ($i_l, i_r$) (red data flow) and between the consecutive frames ($i_l, j_l$) (blue data flow). The computed losses optimize the image-based stereo network. For more details, refer to Sec. \ref{sec:main}.}
	\label{fig:blockdiagram}
\end{figure}
	\begin{itemize}
		\item We develop a test-time training algorithm for consistent video depth estimation, which uses both views of a stereo input. This design resolves the scale ambiguity issues of monocular input. To the best of our knowledge, this work is the first to estimate depth through test-time training on \emph{stereo} video input.
		\item To enforce depth consistency between the stereo views, we propose a new left-right consistency loss, which can be computed time-efficiently.
		\item A gradient-based loss function is introduced to enhance the depth maps by containing more fine-grained detail and sharper depth edges.
		\item The pre-processing time is significantly reduced by exploiting the baseline stereo depth estimator and allowing a faster and simpler algorithm for camera pose computation.
	\end{itemize}
	\section{Related work}
	\textbf{Per-frame Depth Estimation.} This is the most straightforward approach to video depth estimation, which applies one of the image depth estimation algorithms \cite{cheng2020hierarchical,guo2019group,zhang2019ga,Shamsafar_2022_WACV,zhou2017unsupervised,godard2019digging, mde_dataset} frame by frame. The image-based methods can provide a sufficient level of accuracy when applied to each frame of an input video. However, for the whole video sequence, due to independent frame processing, depth may flicker erratically. Random noises and random changes in depth scale are the causes of the flickering effect in depth videos. As a result, per-frame depth estimation is not appropriate for many applications, \eg augmented reality, since inconsistent changes in the background depth break the illusion.
	
	\textbf{Temporally Consistent Methods.} 
	Approaches that account for prior frames are suggested to solve flickering and can be divided into implicit and explicit methods. Implicit algorithms accumulate temporal information in hidden states of a neural network and thus, implicitly exploit the scene geometry. Explicit methods, on the other hand, enforce consistency by gaining an understanding of the scene and camera motion.
	
	As an implicit method, \cite{tananaev2018temporally} presents a simple yet elegant approach that uses a combination of three losses on top of a multi-scale encoder-decoder RNN to use temporal information. Authors of \cite{patil2020don} propose to initialize an RNN with weights of a pre-trained single-frame depth estimation model and fine-tune it on video depth datasets to improve temporal consistency. In \cite{eom2019temporally}, a flow-guided memory unit is proposed as an advancement of a regular GRU unit. By integrating optical flow into an RNN, instead of aggregating features directly after each layer, the authors suggest aligning them based on flow values. Despite the overall improvement in the coherence of depth videos, the training of RNNs is relatively unstable because of the wide range of scenarios, leading to conflicting results in real-world scenes. Also, due to vanishing and exploding gradients, LSTM units cannot aggregate all frames of a video.
	
	In explicit methods \cite{chen2019self, teed2018deepv2d,zhou2017unsupervised,ummenhofer2017demon,vijayanarasimhan2017sfm}, the goal is to estimate camera motion, which in turn is used to re-project pixels between frames and penalize depth differences in corresponding pixels. Hence, these methods can break into the estimation of depth and motion. The output of one task is passed to the other to refine the initial estimates. While implicit methods present simple end-to-end deep models and provide a high FPS rate during inference, explicit algorithms are slower during both training and inference. In terms of quality, explicit methods yield more accurate, detailed, and realistic results.
	
	For training, both implicit and explicit methods usually require large, diverse datasets and do not provide accurate results for scenes that do not match the data distribution of the training dataset. To deal with this limitation, the third class of algorithms was introduced.
	
	\textbf{Test-time Training (TTT).} This class differs from others in that it adapts to each input video independently and does not require a training dataset. In a recent approach \cite{CVDE}, the focus is on geometrical consistency since depth videos, in general, may not be temporally consistent due to camera movement. The authors present a TTT algorithm that fine-tunes the weights of an existing single-frame monocular network to fit the geometrical constraints of a specific video. Also, in this work, the distance of 3D points to the camera depends on the camera pose. Since monocular depth estimators are scale-ambiguous and can only predict the relative positions of objects in a scene correctly, they fail to predict real (metric) depth properly. We address this limitation by developing a test-time training pipeline for stereo video input.
	\section{Methodology}
	\label{sec:main}
	\subsection{Consistent Stereo Video}
	It is clear that in the absence of stereo information, only \emph{relative} depth can be computed. Thus, monocular approaches are scale-ambiguous, \ie depth values for a given 3D point may differ over time due to differences in depth scale. To resolve the underlying scale inconsistency issue, we propose to incorporate stereo input into the pipeline. Given an image pair as input, a stereo-based model estimates \emph{real} depth. More specifically, it minimizes the matching cost of the rectified images, which allows it to compute the accurate disparity map, $d$. Then, metric depth can be computed by $D = UB/d$ from stereo baseline $B$ and focal length $U$. Unlike the monocular depth maps, stereo depth information is consistent with the actual scale of the scene. 
	In contrast to CVDE \cite{CVDE}, we do not have to scale the initial monocular-based depth and make it scale-consistent. CVDE's preprocessing steps, which include the calculation of camera poses and sparse depth maps (assumed to be real depth) by COLMAP \cite{schonberger2016structure}, as well as scale calibration, can therefore be skipped. Moreover, we introduce more geometrical constraints by proposing a loss function that uses depth maps for left-right consistency check.
	
	Figure \ref{fig:blockdiagram} illustrates the block diagram of our proposed method for TTT on stereo sequences. To enforce the geometrical consistency between two frames (pairwise optimization), we consider two sets as $S_{LR} = \bigcup_{0 \le i \le N} (i_l, i_r)$ and $S_{T} = \bigcup_{0 \le i \le N} (i_l, j_l)$. Here, $N$ is the number of timestamps of the stereo video; $i$ and $j$ denote the timestamps, and the subscripts $l$ and $r$ correspond to the left and right images. While $S_{LR}$ considers the left-right frames in each timestamp to enforce the left-right consistency, $S_{T}$ is aimed at the pairwise optimization of frames in the time domain. 
	
	Our formulation requires a matrix, like $Q=[R|T]$, to relate the 3D point coordinates of two frames. Here, $R$ and $T$ represent the rotation matrix and translation vector from one camera's coordinate system to the others. If the two frames belong to a stereo pair, \ie $S_{LR}$, the reprojection matrix can simply be computed since the relative location of the stereo camera lenses is known. Traditionally, stereo lenses are located on the $x$-axis at a distance of the baseline, $B$. For frame pairs across the time domain, \ie $S_{T}$, $Q$ can be computed by a Simultaneous Localization, and Mapping (SLAM) tool like \cite{murORB2}. The 3D point coordinates can then be related by this reprojection matrix as follows:
	\begin{equation}
		\begin{bmatrix}
			x_1 \\ y_1 \\z_1 \\1
		\end{bmatrix}
		= Q * \begin{bmatrix}
			x_2 \\ y_2 \\z_2 \\1
		\end{bmatrix},
		\label{eq:reproj}
	\end{equation}
	where $x_1, y_1, z_1$, and $x_2, y_2, z_2$ denote the 3D points in the camera coordinate systems of the two frames. Note that Eq. \ref{eq:reproj} is only valid for \textit{real} depth, which is not applicable to monocular depth estimation. Therefore, this reprojection theoretically yields incorrect reprojected points in a monocular scenario.
	
	\noindent \textbf{Preprocessing:} We compute optical flow for each pair of the images in $S_{LR}$ and $S_{T}$ and denote the corresponding optical flows as $F_{i_l \rightarrow i_r}$ and $F_{i_l \rightarrow j_l}$, respectively. To account for dis-occlusions between frames, we perform a flow consistency check and produce maps like $M_{i_l \rightarrow i_r}$ and $M_{i_l \rightarrow j_l}$, where pixels are marked as 1 if their forward-backward error exceeds 1 pixel. A joint dis-occluded area is calculated as the intersection of the two maps, $M$. In the following, we describe our test-time training together with the left-right consistency optimization process.
	
	\noindent \textbf{Test-time training:} Given a stereo pair of frames $(i_l, i_r)$ for timestamp $i$, for each pixel $x \in i_l$, we displace its location by optical flow $F_{i_l \rightarrow i_r}(x)$ to compute \textit{flow-displaced} pixel in the right frame $i_r$, as $f_{i_l \rightarrow i_r} = x + F_{i_l \rightarrow i_r}(x)$. The next step is to compute the depth-reprojected point in $i_r$. First, we lift the pixel coordinate $x \in i_l$ to 3D point $c_{i_l}(x)$ in the left frame coordinate system by $c_{i_l}(x) = D_{i_l}(x)K_l^{-1}x$. Here, $K_l$ is the intrinsic parameters of the left lens, and $D_{i_l}$ is the depth estimation for the left image. While in our pipeline, $D_{i_l}$ is estimated by a stereo model, it can also be obtained by monocular methods, like in \cite{CVDE}. The 3D point of $c_{i_l}(x)$ is then projected to the right frame coordinate system using the reprojection matrix. Similarly, we compute the depth-reprojected 3D point across the time domain for the $(i_l, j_l)$ frame pair. Hence, we have:
	\begin{align}
		c_{i_l \rightarrow i_r}(x) = Qc_{i_l},\ \ \ \ \  c_{i_l \rightarrow j_l}(x) = Q_{ij}c_{i_l}
		\label{eq:depth-repr-vcl_t}
	\end{align}
	Note that while the reprojection matrix for the stereo pair is fixed, for the paired left frames across time, \ie $(i_l, j_l)$, $Q_{ij}$ needs to be computed.
	Finally, we compute the depth-reprojected \emph{pixels} $p_{i_l \rightarrow i_r}$ and $p_{i_l \rightarrow j_l}$ by projecting the 3D point of $c_{i_l \rightarrow i_r}(x)$ and $c_{i_l \rightarrow j_l}(x)$ to the coordinate system of $i_r$ and $j_l$, respectively, as follows:
	\begin{equation}
		\begin{split}
			p_{i_l \rightarrow i_r}(x)=\pi(K c_{i_l \rightarrow i_r}(x)), \\
			p_{i_l \rightarrow j_l}(x)=\pi(K c_{i_l \rightarrow j_l}(x)),
		\end{split}
		\label{eq:depth-repr}
	\end{equation}
	with $\pi([x,y,z]^{T})=[x/z,y/z]^{T}$.
	We break our left-right and time consistency losses into two parts: image-space loss \cite{CVDE} and disparity-space loss \cite{CVDE}. The spatial loss is computed as the L2-distance between flow-displaced and depth-reprojected \textit{pixel} coordinates. For the $(i_l, i_r)$ frame pair, it is computed as:
	\begin{align}
		\mathcal{L}_{ i_l \rightarrow i_r }^\textit{spatial}(x)=\| p_{ i_l \rightarrow i_r }(x)-f_{ i_l \rightarrow i_r }(x)\|_2.
	\end{align}
	We compute the disparity loss as L1-distance between \textit{inverse depths} of flow-displaced and depth-reprojected points:
	\begin{align}
		\mathcal{L}_{ i_l \rightarrow i_r  }^\textit{disparity}(x)=u_{i_l}\left|z_{ i_l \rightarrow i_r  }^{-1}(x)-z_{i_r}^{-1}(f_{ i_l \rightarrow i_r  }(x))\right|.
	\end{align}
	Here, $z_{ i_l \rightarrow i_r  } $ and $z_{i_r}$ are the $z$-coordinate of $c_{i_l \rightarrow i_r}$ and $c_{i_r}$ respectively, and $u_{i_l}$ is the focal length of the left lens. 
	
	Likewise, we compute $\mathcal{L}_{ i_l \rightarrow j_l }^\textit{spatial}(x)$ and $\mathcal{L}_{ i_l \rightarrow j_l }^\textit{disparity}(x)$ for the paired left frames, \ie $(i_l, j_l)$.
	
	The final left-right consistency loss for timestamp $i$ of a stereo video is computed as a sum of left-to-right losses over all pixels $x$ with the valid flow:
	\begin{equation}
		\mathcal{L}_{i_l \rightarrow i_r} = \frac{1}{|M|}\sum_{x \in M}\mathcal{L}_{i_l \rightarrow i_r}^\textit{spatial}(x) +
		\lambda \mathcal{L}_{i_l \rightarrow i_r}^\textit{disparity}(x).
		\label{eq:loss_leftright}
	\end{equation}
	For the pixels of image pairs in the time domain, the final loss is as follows:
	\begin{equation}
		\mathcal{L}_{i_l \rightarrow j_l} = \frac{1}{|M|}\sum_{x \in M}\mathcal{L}_{i_l \rightarrow j_l}^\textit{spatial}(x) +
		\lambda \mathcal{L}_{i_l \rightarrow j_l}^\textit{disparity}(x).
		\label{eq:loss_time}
	\end{equation}
	In both Eq. \ref{eq:loss_leftright} and \ref{eq:loss_time}, $\lambda=0.1$. We can then compute the overall geometric consistency loss as a sum of losses over all pairs in $S_{LR}$ and $S_T$:
	\begin{align}
		\mathcal{L}^\textit{geometric} &= \mathcal{L}^\textit{LR} + \mathcal{L}^\textit{T}\\
		&= \sum_{(i_l,i_r) \in S_{LR}} \mathcal{L}_{i_l \rightarrow i_r} + \sum_{(i_l,j_l) \in S_T} \mathcal{L}_{i_l \rightarrow j_l}
		\label{eq:loss_time_final}
	\end{align}
	This loss is backpropagated to the weights of the deep model for depth estimation. With $\mathcal{L}^\textit{geometric}$, the algorithm enforces geometrical consistency both in the time domain and between left and right images.
	\subsubsection{Stereo-based Depth Estimation}
	In contrast to \cite{CVDE}, video stereo input allows us to use stereo-based depth estimation instead of monocular depth estimation. As the backbone stereo network, we use LEAStereo \cite{cheng2020hierarchical} and compute the geometrical consistency loss, $\mathcal{L}^\textit{geometric}$, and backpropagate it through the weights of the stereo model. Because LEAStereo creates a depth map for the left view of stereo input, we only compute the depth-reprojected and flow-displaced points in the TTT pipeline for the \textit{left} view, and the right view is only used to calculate depth. Another important point is that in the stereo model described by \cite{cheng2020hierarchical}, the left and right lenses of a stereo camera must be aligned horizontally, so we use a dataset where both lenses are aligned horizontally.
	\subsubsection{SLAM Integration} The other limitation associated with CVDE \cite{CVDE} is the time complexity. The process of full dense reconstruction in \cite{CVDE}, \ie the Multi-View Stereo (MVS) phase of COLMAP \cite{schonberger2016structure}, is the most time-consuming step and contributes up to 95\% of the total computation time. Besides time issues, COLMAP is prone to error in challenging cases, \eg low-textured regions, insufficient camera translation, etc. Our stereo modification allows replacing COLMAP with another camera motion estimation tool, \eg a Simultaneous Localization, and Mapping (SLAM), provided that it estimates the camera pose on a real scale. While Structure from Motion (SfM) tools, like COLMAP, aim to solve the scene reconstruction problem simultaneously for all observations, SLAM tools solve the problem incrementally. This difference originates from differences in the intended application of SfM and SLAM; SfM was originally designed to reconstruct an entire scene from large \textit{unordered} collections of images, possibly taken with different cameras. On the other hand, SLAM aims to estimate camera pose from a sequence of images taken with a moving agent (\eg a robot). Moreover, SLAM algorithms can track the camera in real time. This helps the preprocessing step of our modified algorithm to take significantly less time compared to the baseline CVDE. Thus, this is a gain from using stereo input, which makes using the SLAM tool possible. For camera pose estimation, we use the stereo regime of a Visual SLAM tool, ORB-SLAM2 \cite{murORB2}. This tool accurately estimates the camera trajectory for the left camera, given a sequence of stereo pairs, the intrinsic camera parameters, and the stereo camera baseline. Also, it is designed to process stereo pairs that are horizontally aligned.
	\subsection{Edge enhancement}
\begin{figure*}[tbp]
	\centering
	\begin{minipage}[t]{\linewidth}
		\centering
		\includegraphics[width=0.32\linewidth]{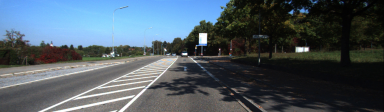}
		\includegraphics[width=0.32\linewidth]{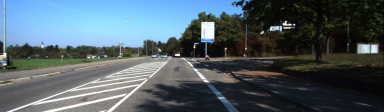}
		\includegraphics[width=0.32\linewidth]{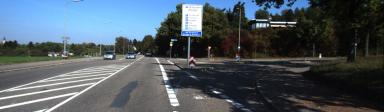}
	\end{minipage}
	\begin{minipage}[t]{\linewidth}
		\centering
		\includegraphics[width=0.32\linewidth]{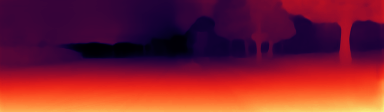}
		\includegraphics[width=0.32\linewidth]{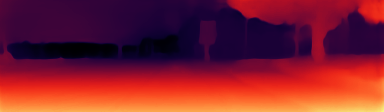}
		\includegraphics[width=0.32\linewidth]{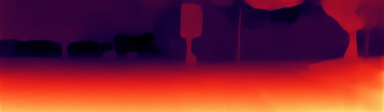}
	\end{minipage}
	\begin{minipage}[t]{\linewidth}
		\centering
		\includegraphics[width=0.32\linewidth]{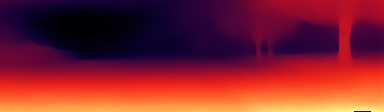}
		\includegraphics[width=0.32\linewidth]{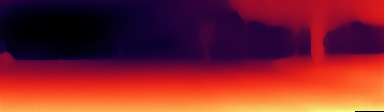}
		\includegraphics[width=0.32\linewidth]{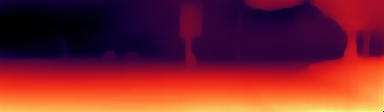}
	\end{minipage}
	\begin{minipage}[t]{\linewidth}
		\centering
		\includegraphics[width=0.32\linewidth]{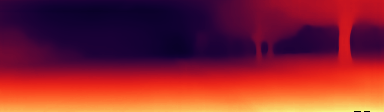}
		\includegraphics[width=0.32\linewidth]{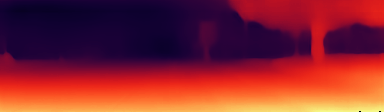}
		\includegraphics[width=0.32\linewidth]{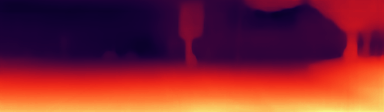}
	\end{minipage}
	\caption{Problem of depth blurriness in test-time training process. The top row shows input color frames at different instances from KITTI odometry \cite{Geiger2012CVPR} dataset. Each consecutive row represents predicted depth maps after 0, 1 and 2 epochs, respectively.}
	\label{fig:edge_loss}
\end{figure*}

	In this section, we address the issue of depth blurriness, which occurs when edges and high-frequency information are lost during the TTT process. As an example of this problem, in Fig. \ref{fig:edge_loss}, the depth map changes with each iteration after 0, 1, and 2 epochs. After the first prediction by a monocular depth estimation \cite{ranftl2019towards}, which contains sharp edges on object borders and small details, depth maps get more blurred as compared to the previous iterations. 
	During the process, we aim to preserve the high-frequency information that is present on the initial depth maps (Fig. \ref{fig:edge_loss}, 2nd row). We achieve this by using gradient-based loss during training between the original depth map and the current depth estimation (at each iteration). The proposed method can be applied to both monocular and stereo-based depth setups. In the following, we describe two gradient-based losses. 
	
	\noindent \textbf{Multi-scale Gradient Loss.} 
	We propose to adopt a multi-scale gradient loss \cite{ummenhofer2017demon} as the scale-invariant gradient loss. This discrete scale-invariant gradient $g$ is defined as:
	\begin{equation}
		\label{eq:graddef}
		\footnotesize
		\begin{split}
			g_h[I](m,n)=(\frac{I(m+h,n)-I(m,n)}{|I(m+h,n)|+|I(m,n)|},\\
			\frac{I(m,n+h)-I(m,n)}{|I(m,n+h)|+|I(m,n)|}).
		\end{split}
	\end{equation}
	Here, $I$ is the input image on which the gradient is calculated. This gradient definition is a normalized difference of neighboring values. Despite its simplicity, the normalization factor makes edge maps scale-independent. With different spacings ($h$), gradients at different scales can be covered. This loss penalizes depth differences between pixels located at a distance of $h$ pixels from each other and, according to \cite{ummenhofer2017demon}, enforces sharp depth discontinuities while simultaneously increasing smoothness in homogeneous regions. 
	
	We further modify the loss function by additionally applying edge masks for each gradient scale so that the loss is only computed in areas with strong gradients. Applying an edge mask is an important step to avoid the case where only gradients corresponding to object boundaries affect the geometrical consistency. Hence, we set a global gradient threshold $\alpha=0.02$, \ie only 2\% of pixels correspond to edges. Note, we compute the threshold $\alpha$ for $h = 1$ only and set thresholds for other gradient scales to be $\alpha_h = \alpha 2^{h-1}$. Let $D_i^0$ be the initial depth map produced by an existing algorithm for timestamp $i$, and $\tilde{D_i}$ the depth map of that frame after $\epsilon$ epochs of the TTT process. Edge masks are computed for each gradient scale $h$ on initial depth maps $D_i^0$ using $\alpha$ and the gradient defined in Eq. \ref{eq:graddef}, as following:
	\begin{equation}
		M^{h}_{D_i^0}(m,n) = 
		\begin{cases}
			1, & \text{if } |g_h[\tilde{D_i}](m,n)| > \alpha \\
			0, & \text{else.}
		\end{cases}
		\label{eq:edgemask}
	\end{equation}
	We then compute our multi-scale gradient loss in areas of the predicted depth map that correspond to edges in the initially predicted depth map:
	\begin{equation}
		\footnotesize
		\mathcal{L}_i^\textit{MS} =
		\sum_{h \in {\{1,2,4,6,8\}}} \sum_{(m,n) \in M^{h}_{D_i^0}} \| g_h[D_i^0](m,n) - g_h[\tilde{D_i}](m,n)\|_2.
		\label{eq:msgl}
	\end{equation}
	\noindent \textbf{Contrastive Loss.}
	The second loss is a contrast-based loss inspired by \cite{RCVDE}. We use a distance measure proposed in \cite{RCVDE} to compute the gradient:
	\begin{equation}
		\footnotesize
		\begin{split}
			g^{r}_h[I](m,n) = (\frac{max(I(m+h, n), I(m,n))}{min(I(m+h, n), I(m,n))}, \\
			\frac{max(I(m, n+h), I(m,n))}{min(I(m, n+h), I(m,n))}  )
		\end{split}
	\end{equation}
	This equation measures the ratio of depth values between neighboring pixels and, thus, is independent of the depth scale. Similar to the multi-scale gradient loss, we compute this ratio-based gradient on 5 different scales, and for each scale, we compute an edge mask. Additionally, we notice that aligning the gradient intensity of predicted depth maps to the gradient intensity of initial depth maps is, in fact, not necessary. Our initial requirement is to retain edge information from depth maps before test-time training; therefore, we only need to ensure that an edge remains present at the same location. Consequently, we optimize a squared difference between the predicted gradient ratio $g^{r}_h[\tilde{D_i}]$ and our edge threshold $\alpha=1.05*2^{h-1}$ for pixels in which an edge is present on the initially-computed depth map $D_i^0$: 
	\begin{equation}
		\footnotesize
		\mathcal{L}_{i}^\textit{C}(m, n) = 
		\begin{cases}
			(\alpha -  g^{r}_h[\tilde{D_i}](m,n))^2, &\text{if } g^{r}_h[D_i^0(m,n)])  > \alpha \\
			0, & \text{else.}
		\end{cases}
		\label{eq:cl}
	\end{equation}
	The overall contrastive loss is computed as a mean of Eq. \ref{eq:cl} for all pixels of any given depth map of height $H$ and width $W$ as follows:
	\begin{equation}
		\footnotesize
		\mathcal{L}_{i}^\textit{C}= \frac{1}{H \times W} \sum_{m,n}  \mathcal{L}_{i}^\textit{C}(m, n).
	\end{equation}
	The above loss is then added to Eq. \ref{eq:loss_time_final} and is backpropagated to the weights of the deep model for depth estimation.
	\section{Experimental results and discussion}
	\subsection{Datasets}
	\noindent\textbf{ETH3D.} To assess our edge enhancement method and consistency loss, we use the ETH3D SLAM dataset \cite{Schops_2019_CVPR}.  The benchmark consists of two-image sequences for each lens of a stereo camera along with ground-truth depths for the left view. We evaluate only one of the views. Although ETH3D stereo camera lenses are not located on the same horizontal axis, this does not pose a problem as a reprojection matrix is provided. 
	
	As an evaluation metric, we use the mean absolute relative error in depth space, which calculates the depth error relative to the ground-truth. Additionally, we compute the percentage of pixels with a relative error exceeding $1.25$, $1.25^2$, and $1.25^3$. Since ETH3D is a dataset collected using DSLR imaging, depth maps with missing values are set to zero and in our evaluation process, we mask out pixels with missing depth values.
	
	\noindent\textbf{KITTI.} To evaluate our Consistent Stereo Video (CSV) model, considering the requirements of LEAStereo and ORB-SLAM2, we need stereo video frames that are aligned horizontally, which ETH3D SLAM does not offer. Therefore, we test our algorithm on the KITTI Odometry/SLAM benchmark \cite{Geiger2012CVPR}, which suits both the stereo depth model and ORB-SLAM2. By using the LEAStereo model \cite{cheng2020hierarchical}, pre-trained on the KITTI Stereo benchmark, and testing on the KITTI Odometry benchmark, we ensure that:
	\textit{i)} Training and test data lie in similar domains, \ie the images are shot with the same stereo camera; \textit{ii)} The evaluation is fair as the stereo model has not been trained on the KITTI Odometry benchmark. We use the 11 sequences of the KITTI Odometry test split to evaluate our algorithm and compare it to the stereo model results.
	
	\noindent\textbf{New Metric.} As the KITTI benchmarks do not provide ground-truth \emph{depth} data, we developed an evaluation metric that does not rely on depth comparison. For timestamp $i$, we estimate the real depth of the left image, $D_{i_l}$. Then, we reproject pixels from the left stream into the right one to get a reprojected image $I^i_{l \rightarrow r}$:
	\begin{equation}
		I^i_{l \rightarrow r} = I^i_{r} \langle proj(D_{i_l}, K, Q) \rangle
	\end{equation}
	Here, $proj(.)$ are the resulting 2D pixel coordinates of the projected depths $D_{i_l}$ in the right camera coordinate system, $K$ is the shared intrinsic parameters, $Q$ is the reprojection matrix, and $\langle . \rangle$ is the sampling operator. 
	For $proj(.)$, we compute depth-reprojected coordinates for each pixel of $I^i_{l}$. By measuring the photometric distance between the real right image and the reprojected one, we can estimate the depth accuracy by:
	\begin{equation}
		L^{photo}_i = dist( I^i_{r}, I^i_{l \rightarrow r}).
		\label{eq:photometric}
	\end{equation}
	$dist()$ is a distance measure in pixel space, for which we use L1- and L2- distances. The final photometric losses are the average of losses in all timestamps.
	\begin{table}[h]
	\begin{center}
		\footnotesize
		\begin{tabular}{lcccc}
			\toprule
			Model & $L_1^{Rel} \downarrow$ & $\delta_{1.25} \downarrow$ &  $\delta^2_{1.25} \downarrow$ & $\delta^3_{1.25} \downarrow$ \\
			\midrule
			MiDaS \cite{ranftl2019towards} &  0.2339 & 0.3254 & 0.1410 &  0.0779 \\
			CVDE \cite{CVDE} & 0.2171 & 0.2730 & 0.0960 & 0.0507 \\
			\midrule
			CVDE + $\mathcal{L}^\textit{MS}$ &  0.2165 & 0.2807 & \textbf{0.0921} & 0.0435 \\
			CVDE + $\mathcal{L}^\textit{C}$ &  \underline{\textbf{0.2143}} &  \underline{\textbf{0.2692}} &  \underline{0.0929} & \underline{0.0434}\\
			\bottomrule
		\end{tabular}
	\end{center}
	\caption{Quantitative evaluation of the proposed gradient-based losses on the ETH3D SLAM dataset. Our multi-scale gradient loss, $\mathcal{L}^\textit{MS}$, and the contrastive loss, $\mathcal{L}^\textit{C}$, outperform MiDaS and CVDE in $L_1^{Rel}$ and $\delta^2_{1.25}$ and in all metrics, respectively.}
	\label{tab:grad_eth}
\end{table}
	\subsection{Edge enhancement}
	We integrate the proposed gradient losses into the CVDE pipeline to evaluate their effectiveness. A comparison of these frameworks with the CVDE baseline and MiDaS v.2 \cite{ranftl2019towards} that was used as the backbone of CVDE, is presented in Tab. \ref{tab:grad_eth}. Since MiDas depth are not in metric space, we apply a similar least-square-based post-processing technique on the MiDas depth as done in the CVDE \cite{CVDE} for comparison and analysis.
	
\begin{figure*}[htpb]
	\begin{center}
		\begin{minipage}[c]{1\linewidth}
			\centering
			\includegraphics[width=0.242\linewidth]{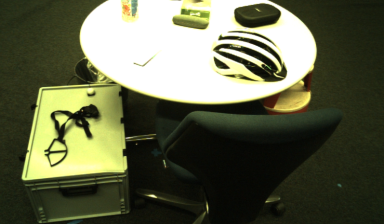}
            \includegraphics[width=0.242\linewidth]{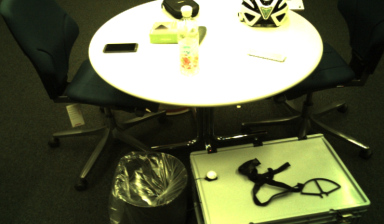}
            \includegraphics[width=0.242\linewidth]{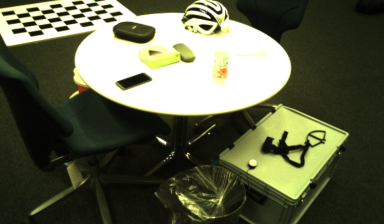}
            \includegraphics[width=0.242\linewidth]{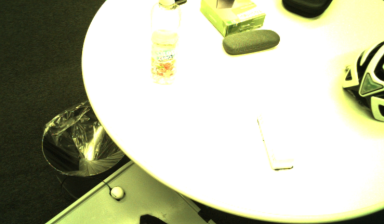}
		\end{minipage}
		\\
		\begin{minipage}[c]{1\linewidth}
			\centering
			\begin{turn}{0}\footnotesize{RGB images}\end{turn}			
		\end{minipage}	
		\\
		\begin{minipage}[c]{1\linewidth}
			\centering
			\includegraphics[width=0.242\linewidth]{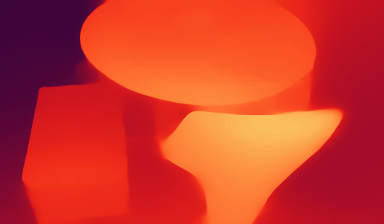}
            \includegraphics[width=0.242\linewidth]{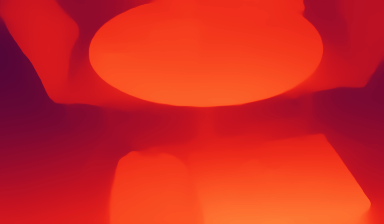}
            \includegraphics[width=0.242\linewidth]{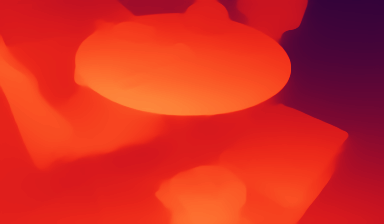}
            \includegraphics[width=0.242\linewidth]{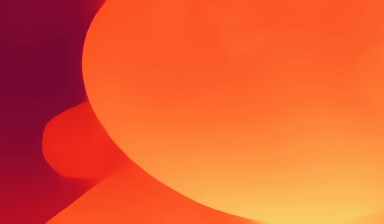}
		\end{minipage}
		\\
		\begin{minipage}[c]{1\linewidth}
			\centering
			\begin{turn}{0}\footnotesize{MiDaS \cite{ranftl2019towards}}\end{turn}			
		\end{minipage}	
		\\
		\begin{minipage}[c]{1\linewidth}
			\centering
			\includegraphics[width=0.242\linewidth]{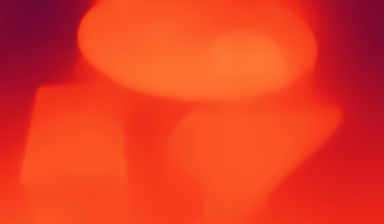}
            \includegraphics[width=0.242\linewidth]{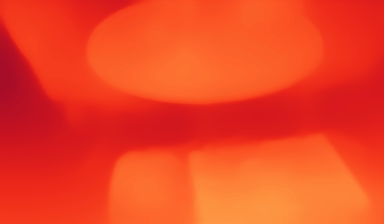}
            \includegraphics[width=0.242\linewidth]{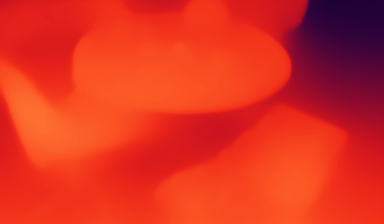}
            \includegraphics[width=0.242\linewidth]{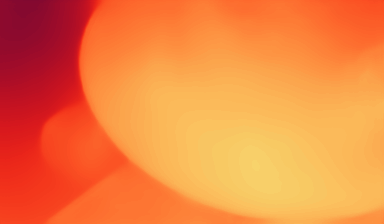}
		\end{minipage}
		\\
		\begin{minipage}[c]{1\linewidth}
			\centering
			\begin{turn}{0}\footnotesize{CVDE \cite{CVDE}}\end{turn}			
		\end{minipage}	
		\\
		\begin{minipage}[c]{1\linewidth}
			\centering
			\includegraphics[width=0.242\linewidth]{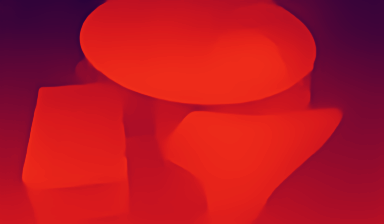}
            \includegraphics[width=0.242\linewidth]{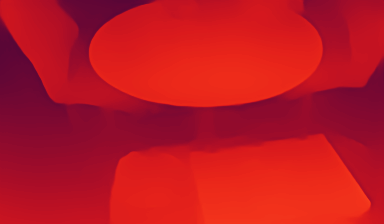}
            \includegraphics[width=0.242\linewidth]{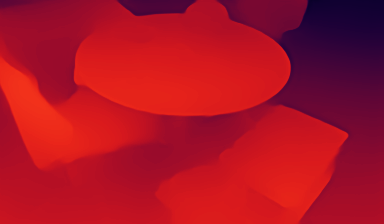}
            \includegraphics[width=0.242\linewidth]{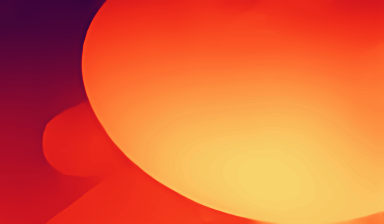}
		\end{minipage}
		\\
		\begin{minipage}[c]{1\linewidth}
			\centering
			\begin{turn}{0}\footnotesize{CVDE + Multi-scale Gradient Loss}\end{turn}			
		\end{minipage}	
		\\
		\begin{minipage}[c]{1\linewidth}
			\centering
			\includegraphics[width=0.242\linewidth]{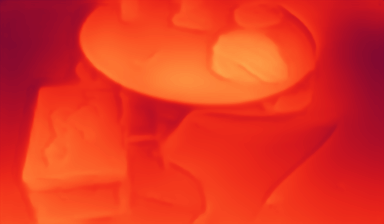}
            \includegraphics[width=0.242\linewidth]{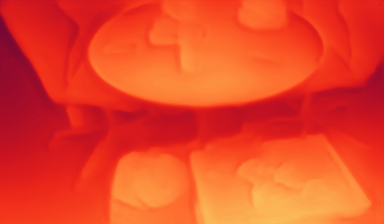}
            \includegraphics[width=0.242\linewidth]{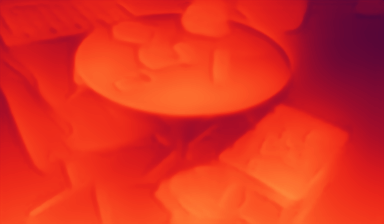}
            \includegraphics[width=0.242\linewidth]{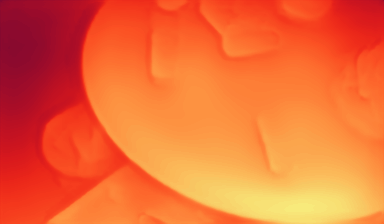}
		\end{minipage}
		\\
		\begin{minipage}[c]{1\linewidth}
			\centering
			\begin{turn}{0}\footnotesize{CVDE + Contrastive Loss}\end{turn}			
		\end{minipage}	
		\\
		\caption{Sample depth maps on ETH3D dataset. Note how our proposed edge enhancement method preserves the details and creates sharper depth maps.}
		\label{fig:eth3d_grad_vis}
	\end{center}
\end{figure*}

	MiDaS has the highest error in all metrics according to Tab. \ref{tab:grad_eth}, confirming that frame-by-frame depth estimation is causing depth inconsistencies. We can see that our gradient-based losses improve the results achieved by CVDE. While the multi-scale gradient loss outperforms both MiDaS and CVDE in $L_1^{Rel}$ and $\delta^2_{1.25}$ metrics, the contrastive loss surpasses MiDaS and CVDE in all the metrics. Figure \ref{fig:eth3d_grad_vis} presents a visualization of depth maps on ETH3D image sequences. CVDE with multi-scale gradient loss in these samples slightly improves over the original CVDE results. On the other hand, CVDE with contrastive loss produces the most detailed depth maps. Hence, both the quantitative and qualitative evaluations indicate that introducing edge data by the proposed contrastive loss achieves significant improvement over the original CVDE loss.
	\begin{table}[h]
	\begin{center}
		\footnotesize
		\begin{tabular}{lcccc}
			\toprule
			Model & $L_1^{Rel} \downarrow$ & $\delta_{1.25} \downarrow$ &  $\delta^2_{1.25} \downarrow$ & $\delta^3_{1.25} \downarrow$\\
			\midrule
			MiDaS \cite{ranftl2019towards} &  0.2339 & 0.3254 & 0.1410 &  0.0779 \\
			Ours, $\mathcal{L}^\textit{LR}$ (\emph{mono})  & 0.2302 &  0.2956 & 0.1264 & 0.0634 \\
			\midrule
			CVDE \cite{CVDE} (\emph{mono}) & 0.2171 & 0.2730 & 0.0960 & 0.0507 \\
			Ours, $\mathcal{L}^\textit{LR}$, $\mathcal{L}^\textit{T}$ (\emph{mono}) & \textbf{0.2151} & \textbf{0.2520}  & \textbf{0.0920} & \textbf{0.0491} \\
			\bottomrule
		\end{tabular}
	\end{center}
	\caption{Quantitative evaluation of our proposed left-right consistency loss on the ETH3D dataset. Note that MiDaS and \quotes{Ours, $\mathcal{L}^\textit{LR}$} use per-frame estimation without applying geometric consistency across time. \quotes{mono} denote fine-tuning the monocular model, \ie MiDaS in the TTT process.}
	\label{tab:vcl_eval}
\end{table}
	\begin{table}[h]
	\centering
	\footnotesize
	\begin{tabular}{lcc}
		\toprule
		Model & $L_1^{photo} \downarrow$ & $L_2^{photo} \downarrow$ \\
		\midrule
		LEAStereo \cite{cheng2020hierarchical} &  0.1989 & 0.0985  \\
		CVDE \cite{CVDE} & 0.1850  & 0.0954  \\
		\midrule
		Ours, $\mathcal{L}^\textit{LR}$ & 0.1721 & 0.0889 \\
		Ours, $\mathcal{L}^\textit{T}$  & 0.1350 & 0.0617 \\
		Ours, $\mathcal{L}^\textit{T}$, $\mathcal{L}^\textit{C}$ & 0.1214 & 0.0514 \\
		Ours, $\mathcal{L}^\textit{T}$, $\mathcal{L}^\textit{C}$, $\mathcal{L}^\textit{LR}$: \textbf{ECSV}  & \textbf{0.1010}  & \textbf{0.0440} \\
		\bottomrule
	\end{tabular}
	\caption{Quantitative evaluation on the KITTI odometry dataset. LEAStereo \cite{cheng2020hierarchical} is an image-based stereo model. CVDE \cite{CVDE} and our models are TTT approaches. Note that we use the losses to fine-tune the \emph{stereo} model \cite{cheng2020hierarchical}, LEAStereo. $L_1^{photo}$ and $L_2^{photo}$ are computed based on eq. \ref{eq:photometric}. Our comprehensive model (ECSV) outperforms CVDE by a large margin.}
	\label{tab:stereo_eval}
\end{table}

	\begin{figure*}[h]
	\centering
	\begin{minipage}[c]{1\linewidth}
		\centering
		\includegraphics[width=0.242\linewidth]{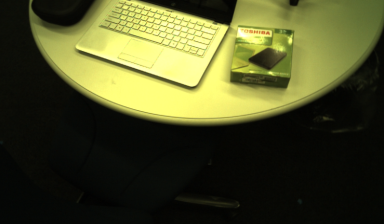}
		\includegraphics[width=0.242\linewidth]{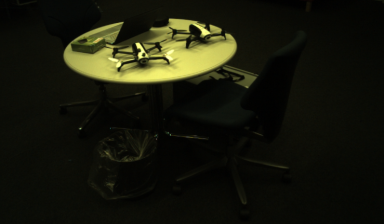}
		\includegraphics[width=0.242\linewidth]{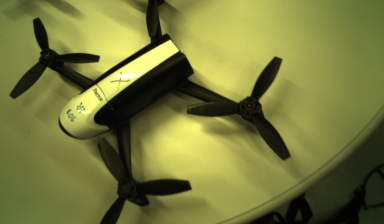}
		\includegraphics[width=0.242\linewidth]{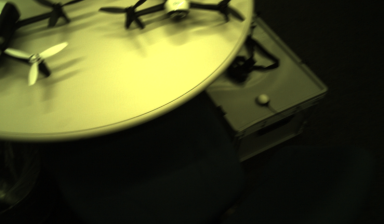}
	\end{minipage}
	\\
	\begin{minipage}[c]{1\linewidth}
		\centering
		\begin{turn}{0}\footnotesize{RGB images}\end{turn}			
	\end{minipage}	
	\\
	\begin{minipage}[c]{1\linewidth}
		\centering
		\includegraphics[width=0.242\linewidth]{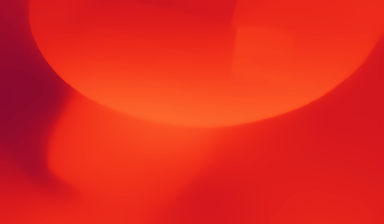}
		\includegraphics[width=0.242\linewidth]{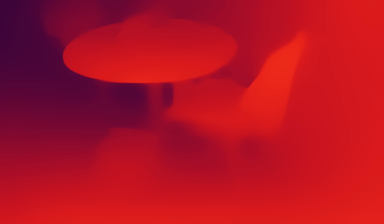}
		\includegraphics[width=0.242\linewidth]{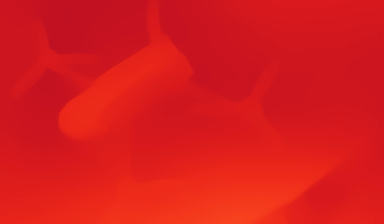}
		\includegraphics[width=0.242\linewidth]{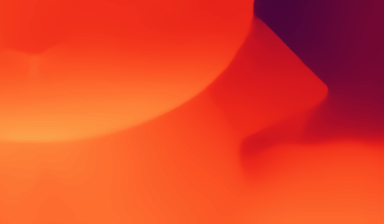}
	\end{minipage}
	\\
	\begin{minipage}[c]{1\linewidth}
		\centering
		\begin{turn}{0}\footnotesize{MiDaS \cite{ranftl2019towards}}\end{turn}			
	\end{minipage}	
	\\
	\begin{minipage}[c]{1\linewidth}
		\centering
		\includegraphics[width=0.242\linewidth]{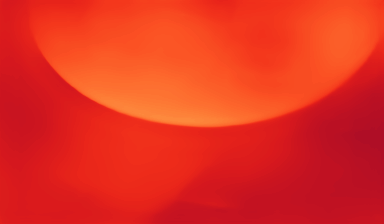}
		\includegraphics[width=0.242\linewidth]{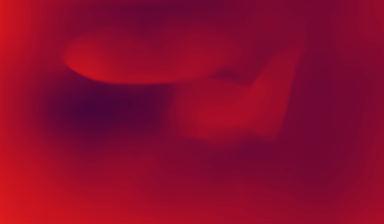}
		\includegraphics[width=0.242\linewidth]{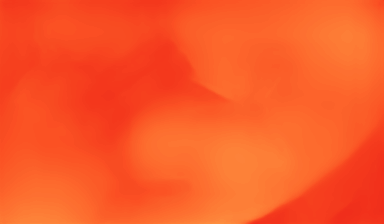}
		\includegraphics[width=0.242\linewidth]{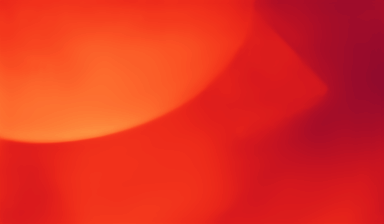}
	\end{minipage}
	\\
	\begin{minipage}[c]{1\linewidth}
		\centering
		\begin{turn}{0}\footnotesize{CVDE \cite{CVDE}}\end{turn}			
	\end{minipage}	
	\\
	\begin{minipage}[c]{1\linewidth}
		\centering
		\includegraphics[width=0.242\linewidth]{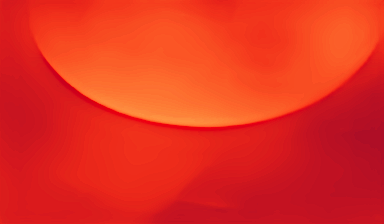}
		\includegraphics[width=0.242\linewidth]{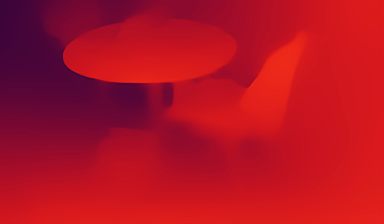}
		\includegraphics[width=0.242\linewidth]{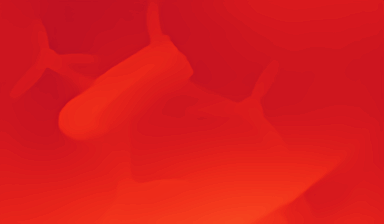}
		\includegraphics[width=0.242\linewidth]{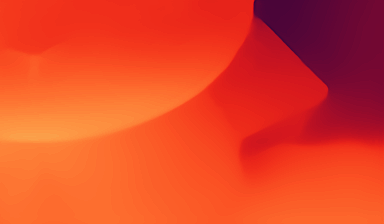}
	\end{minipage}
	\\
	\begin{minipage}[c]{1\linewidth}
		\centering
		\begin{turn}{0}\footnotesize{Ours, $\mathcal{L}^\textit{LR}$, $\mathcal{L}^\textit{T}$}\end{turn}			
	\end{minipage}	
	\\
	\caption{Sample depth maps on ETH3D dataset predicted in rows respectively by: MiDaS, and MiDaS after test-time training with CVDE, with our left-right consistency loss ($\mathcal{L}^\textit{LR}$), and with both $\mathcal{L}^\textit{LR}$ and $\mathcal{L}^\textit{T}$.}
	\label{fig:eth3d_vcl_vis}
\end{figure*}
	\begin{figure*}[htpb]
	\begin{center}
		\begin{minipage}[c]{1\linewidth}
			\centering
			\includegraphics[width=0.32\linewidth]{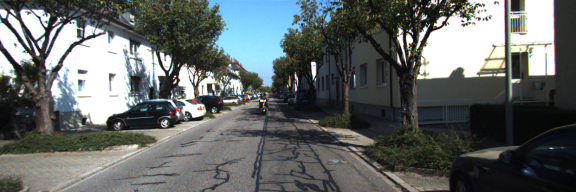}
			\includegraphics[width=0.32\linewidth]{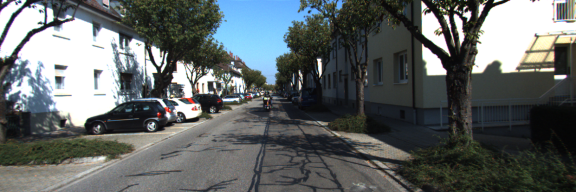}
			\includegraphics[width=0.32\linewidth]{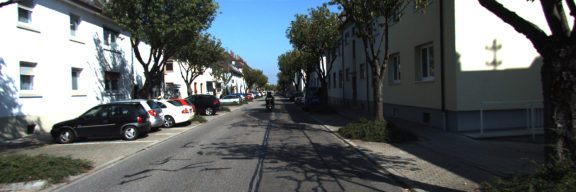}
		\end{minipage}
		\\
		\begin{minipage}[c]{1\linewidth}
			\centering
			\begin{turn}{0}\footnotesize{RGB images}\end{turn}			
		\end{minipage}	
		\\
		\begin{minipage}[c]{1\linewidth}
			\centering
			\includegraphics[width=0.32\linewidth]{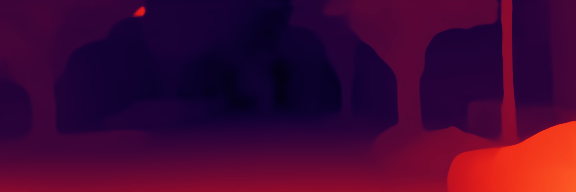}
			\includegraphics[width=0.32\linewidth]{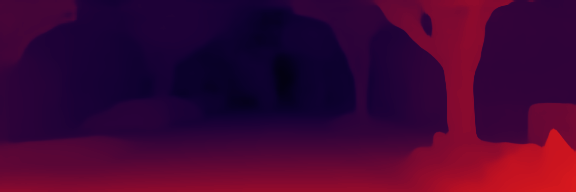}
			\includegraphics[width=0.32\linewidth]{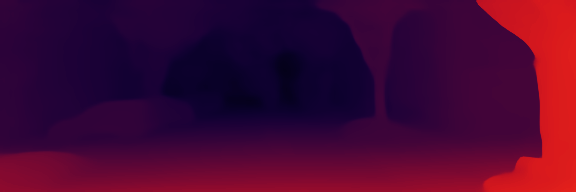}
		\end{minipage}
		\\
		\begin{minipage}[c]{1\linewidth}
			\centering
			\begin{turn}{0}\footnotesize{LEAStereo \cite{cheng2020hierarchical}}\end{turn}			
		\end{minipage}	
		\\
		\begin{minipage}[c]{1\linewidth}
			\centering
			\includegraphics[width=0.32\linewidth]{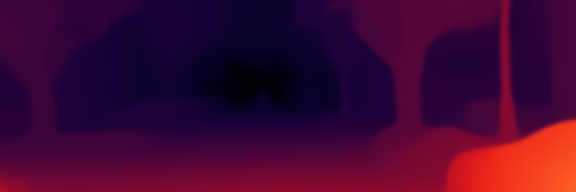}
			\includegraphics[width=0.32\linewidth]{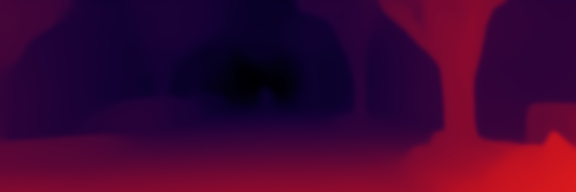}
			\includegraphics[width=0.32\linewidth]{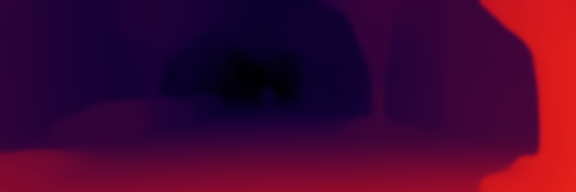}
		\end{minipage}
		\\
		\begin{minipage}[c]{1\linewidth}
			\centering
			\begin{turn}{0}\footnotesize{CVDE \cite{CVDE}}\end{turn}			
		\end{minipage}	
		\\
		\begin{minipage}[c]{1\linewidth}
			\centering
			\includegraphics[width=0.32\linewidth]{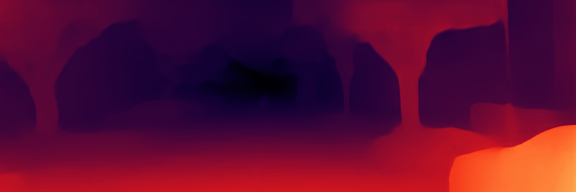}
			\includegraphics[width=0.32\linewidth]{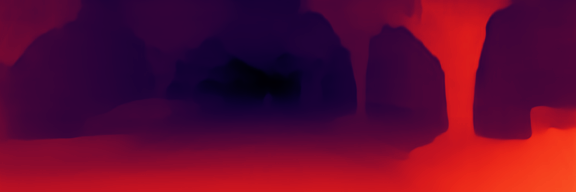}
			\includegraphics[width=0.32\linewidth]{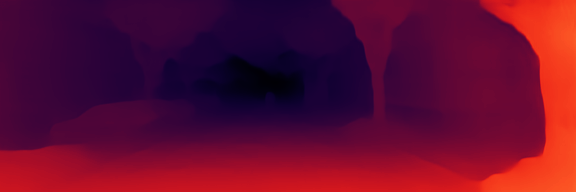}
		\end{minipage}
		\\
		\begin{minipage}[c]{1\linewidth}
			\centering
			\begin{turn}{0}\footnotesize{Ours, $\mathcal{L}^\textit{LR}$}\end{turn}			
		\end{minipage}	
		\\
		\begin{minipage}[c]{1\linewidth}
			\centering
			\includegraphics[width=0.32\linewidth]{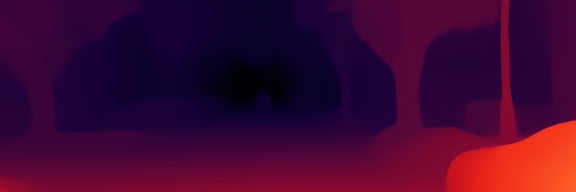}
			\includegraphics[width=0.32\linewidth]{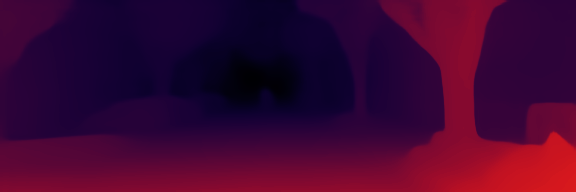}
			\includegraphics[width=0.32\linewidth]{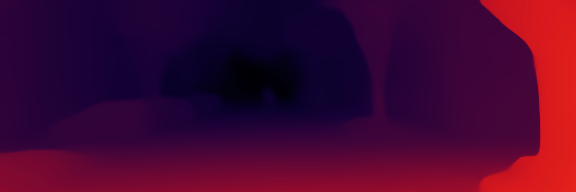}
		\end{minipage}
		\\
		\begin{minipage}[c]{1\linewidth}
			\centering
			\begin{turn}{0}\footnotesize{Ours, $\mathcal{L}^\textit{T}$}\end{turn}			
		\end{minipage}	
		\\
		\begin{minipage}[c]{1\linewidth}
			\centering
			\includegraphics[width=0.32\linewidth]{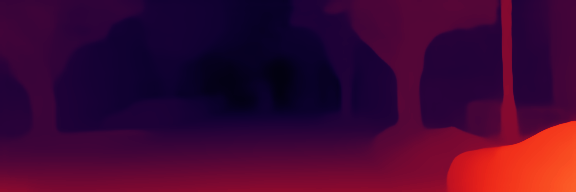}
			\includegraphics[width=0.32\linewidth]{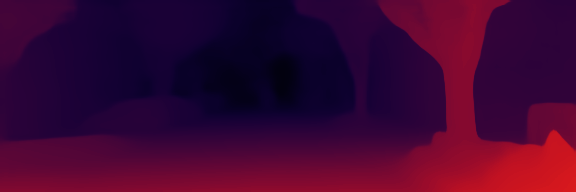}
			\includegraphics[width=0.32\linewidth]{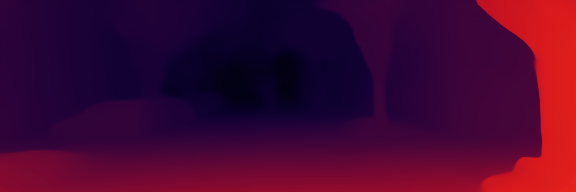}
		\end{minipage}
		\\
		\begin{minipage}[c]{1\linewidth}
			\centering
			\begin{turn}{0}\footnotesize{Ours, $\mathcal{L}^\textit{T}$, $\mathcal{L}^\textit{C}$ }\end{turn}			
		\end{minipage}	
		\\
		\begin{minipage}[c]{1\linewidth}
			\centering
			\includegraphics[width=0.32\linewidth]{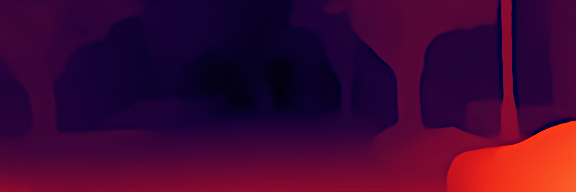}
			\includegraphics[width=0.32\linewidth]{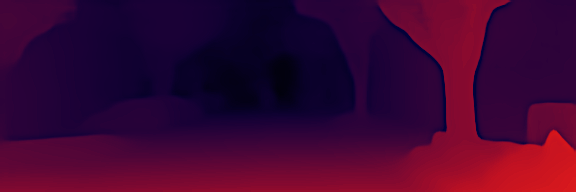}
			\includegraphics[width=0.32\linewidth]{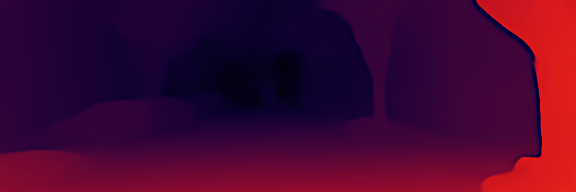}
		\end{minipage}
		\\
		\begin{minipage}[c]{1\linewidth}
			\centering
			\begin{turn}{0}\footnotesize{\textbf{ECSV}: Ours, $\mathcal{L}^\textit{T}$, $\mathcal{L}^\textit{C}$, $\mathcal{L}^\textit{LR}$}\end{turn}			
		\end{minipage}	
		\\
		\caption{Visualization of depth maps on the KITTI odometry dataset for sample consecutive frames. The last row show our the estimated depth maps by our \textbf{E}dge-aware \textbf{C}onsistent \textbf{S}tereo \textbf{V}ideo (\textbf{ECSV}) model.}
		\label{fig:stereo_vis}
		\vspace{-0.5 cm}
	\end{center}
\end{figure*}
	\subsection{Left-right consistency loss}
	To prove the effectiveness of our left-right consistency loss ($\mathcal{L}^\textit{LR}$), we perform our proposed TTT with $\mathcal{L}^\textit{LR}$ and fine-tune the weights of the monocular depth model of MiDaS for 20 epochs, setting the learning rate to 0.0004 with the ADAM optimizer. First, we align predicted and ground-truth depths per frame. Next, we compute the evaluation metrics and calculate the mean of all metrics across 11 sequences of the ETH3D SLAM dataset. Table \ref{tab:vcl_eval} tabulates the performance comparison on a test split of the ETH3D SLAM dataset. In \quotes{MiDaS} and \quotes{Ours, $\mathcal{L}^\textit{LR}$} no temporal information has been accounted for. The $\mathcal{L}^\textit{LR}$ loss surpasses MiDaS, proving stereo consistency's benefit. When using temporal consistency, our framework outperforms the original CVDE model. In Fig. \ref{fig:eth3d_vcl_vis}, we have shown depth maps on one of the image sequences of the ETH3D dataset. Our pipeline qualitatively improves over the CVDE baseline.
	\subsection{Consistent Stereo Video (CSV)}
	We make several adjustments to evaluate our stereo-assisted TTT on the KITTI Odometry dataset. Firstly, we replace hierarchical frame sampling in CVDE with consecutive cases: $S-T = \bigcup_{i=1}^{ i=n} (i-1, i) $, due to the fast-forward motion of the camera in this dataset and insufficient overlap between distant frames. For optical flow computation, we use the FlowNet2 model \cite{IMKDB17}, fine-tuned on a combination of KITTI2012 and KITTI2015 \cite{Menze2015ISA} datasets. The input images are down-sampled to $384\times112$ resolution, and the TTT procedure runs for 20 epochs, with a learning rate of $4\times10^{-5}$ and the ADAM optimizer. For comparative evaluation, Tab. \ref{tab:stereo_eval} presents the average L1- and L2-photometric errors for the following algorithms in order: LEAStereo \cite{cheng2020hierarchical}, CVDE \cite{CVDE}, \quotes{Ours, init with LEAStereo, $\mathcal{L}^\textit{LR}$, $\mathcal{L}^\textit{T}$}, \quotes{Ours, init with LEAStereo, $\mathcal{L}^\textit{LR}$, $\mathcal{L}^\textit{T}$}. According to our metrics, the TTT procedure significantly improves the results of the image-based stereo model, \ie LEAStereo. Additionally, our model with contrastive loss provides a slight improvement when optimized jointly with the geometrical consistency loss during TTT. As expected, this approach surpasses the monocular-based CVDE algorithm by a large margin. Figure \ref{fig:stereo_vis} presents a visualization of depth maps produced by each algorithm in consideration. Both the quantitative evaluation and visualization confirm that the fine-tuning weights of LEAStereo following the TTT pipeline with left-right and time geometry consistency improves the results.
	\subsection{Time complexity}
	We analyze the computation time in Table \ref{tab:time_vcl} on a sub-sequence of 400 images from the KITTI Odometry dataset. As can be seen, the proposed left-right consistency loss requires less time than the original CVDE. This is because the proposed $\mathcal{L}^\textit{LR}$ loss relies only on the relative poses of the stereo views and does not require camera pose computation. The original CVDE runs a costly dense COLMAP reconstruction to estimate camera poses and calibrate depth scale. Furthermore, we compare the timing of the original CVDE algorithm and the proposed \emph{stereo-based} TTT. Our comprehensive algorithm, ECSV, is faster than CVDE due to the time difference of ORB-SLAM2 camera pose computation.
	\begin{table}[h]
	\footnotesize
	\centering
	\begin{tabular}{lc}
		\toprule
		Model & Time ($s$), on KITTI\\
		\midrule
		CVDE \cite{CVDE} &   6936 (\emph{mono})\\
		\midrule
		CVDE \cite{CVDE} + $\mathcal{L}^\textit{C}$ &  6949 (\emph{mono}) \\
		Ours, $\mathcal{L}^\textit{T}$ &   2850 (\emph{stereo}) \\
		Ours, $\mathcal{L}^\textit{LR}$ & \textbf{1200} (\emph{stereo})\\
		Ours, $\mathcal{L}^\textit{T}$, $\mathcal{L}^\textit{C}$, $\mathcal{L}^\textit{LR}$: \textbf{ECSV} & \underline{3180} (\emph{stereo}) \\
		\bottomrule
	\end{tabular}
	\caption{Timing comparison. \quotes{mono} and \quotes{stereo} denote fine-tuning the monocular (MiDaS) and stereo (LEAStereo) models in the TTT process, respectively. Compared to the mono-based TTT method, \ie CVDE, our stereo-based model, ECSV, requires less computation time. }
	\label{tab:time_vcl}
\end{table}
	\subsection{SLAM Integration.}
\begin{figure*}[h]
	\centering
	\includegraphics[width=0.495\linewidth]{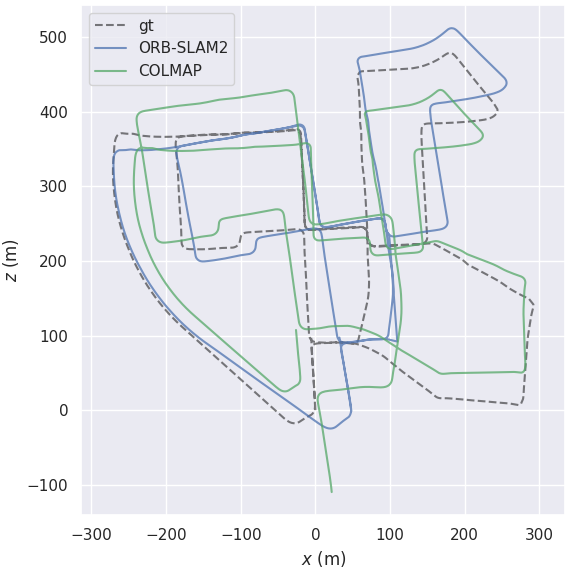}
	\includegraphics[width=0.495\linewidth]{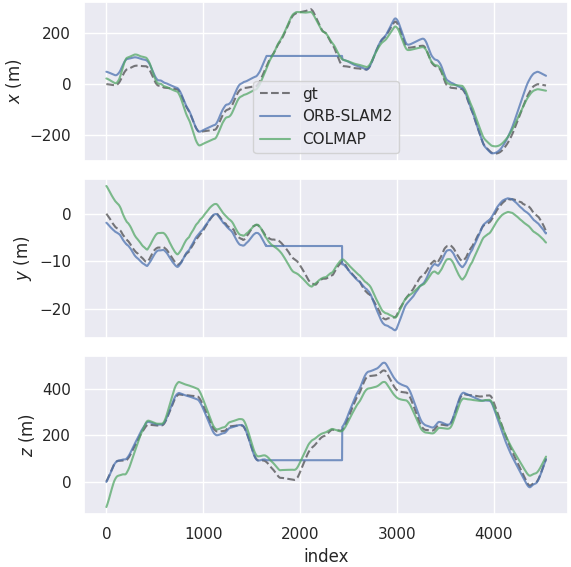}
	\caption{Trajectories of COLMAP vs. ORB-SLAM2 for a sequence of KITTI dataset. The projections of trajectories on $xz$-plane (\emph{Left}) and on $x$-, $y$- and $z$-axis (\emph{Right}).}
	\label{fig:traj_comparison}
\end{figure*}

	In order to validate the benefits of using SLAM, which is made possible by using stereo input, we compare the performance of COLMAP \cite{schonberger2016structure} and ORB-SLAM2 \cite{murORB2}. The KITTI Odometry dataset is used to compute camera poses based on the standard settings of both algorithms. Also, Fig. \ref{fig:traj_comparison} shows plots of trajectories against the ground-truth for a KITTI Odometry sequence with 300 images. Compared to COLMAP, ORB-SLAM provides more accurate camera trajectories. COLMAP and ORB-SLAM2 accuracy and computation time are compared in Tab. \ref{tab:colmap-orb}.
	\begin{table}[h]
	\centering
	\resizebox{\columnwidth}{!}{
		\begin{tabular}{lcccc}
			\toprule
			& Time ($s$) & Mean error ($m$) & Median error ($m$) & RMSE ($m$)  \\
			\midrule
			COLMAP & 4762 & 169.9 & 173.33 & 190.09\\
			ORB-SLAM2 & \textbf{180} & \textbf{45.02} & \textbf{27.73} & \textbf{64.95}\\
			\bottomrule
	\end{tabular}}
	\caption{COLMAP vs. ORB-SLAM2 in accuracy and time.}
	\label{tab:colmap-orb}
\end{table}
	\section{Conclusion}
	We presented an algorithm for consistent video depth estimation, and unlike the existing methods, we used stereo video sequences as the input. This way, we can enforce geometric consistency between the left and right images and across the time domain. We also proposed a gradient loss function to address the problem of blurriness in the test-time training methods. The quantitative and qualitative analyses validate our framework, which also demonstrates faster inference by getting assistance using stereo data.
	{\small
		\bibliographystyle{ieee_fullname}
		\bibliography{egbib}
	}
\end{document}